%% file: main.tex
\documentclass{article}
\usepackage{spconf,amsmath,epsfig}

\title{AUTHOR GUIDELINES FOR IGARSS 2020 MANUSCRIPTS}
\name{Author(s) Name(s)\thanks{Thanks to XYZ agency for funding.}}
\address{Author Affiliation(s)}
\usepackage{spconf,amsmath,graphicx}
\usepackage{times}
\usepackage{epsfig}
\usepackage{amssymb}
\usepackage{algorithm}
\usepackage{multirow}
\usepackage{multicol}
\usepackage[noend]{algpseudocode}
\algnewcommand{\LeftComment}[1]{\Statex \(\triangleright\) #1}
\usepackage[dvipsnames]{xcolor}

\usepackage[pagebackref=true,breaklinks=true,colorlinks,bookmarks=false]{hyperref}
\newcommand{\figref}[1]{Figure~\ref{fig:#1}}
\newcommand{\tblref}[1]{Table~\ref{tab:#1}}
\newcommand{\eqnref}[1]{Eqn.~\ref{eqn:#1}}

\title{Intensity Harmonization for Airborne LiDAR}
\name{David Jones, Nathan Jacobs
}
\address{Department of Computer Science, University of Kentucky}
\begin{document}
%
\maketitle
\begin{abstract}
    Constructing a point cloud for a large geographic region, such as a state or country, can require multiple years of effort. Often several vendors will be used to acquire LiDAR data, and a single region may be captured by multiple LiDAR scans. A key challenge is maintaining consistency between these scans, which includes point density, number of returns, and intensity. Intensity in particular can be very different between scans, even in areas that are overlapping. Harmonizing the intensity between scans to remove these discrepancies is expensive and time consuming. In this paper, we propose a novel method for point cloud harmonization based on deep neural networks. We evaluate our method quantitatively and qualitatively using a high quality real world LiDAR dataset. We compare our method to several baselines, including standard interpolation methods as well as histogram matching. We show that our method performs as well as the best baseline in areas with similar intensity distributions, and outperforms all baselines in areas with different intensity distributions. Source code is available at \url{https://github.com/mvrl/lidar-harmonization}.
    
\end{abstract}

\input{sections/1intro}

\input{sections/2related_works}
\input{sections/3problem_def}

\input{sections/5evaluation}

\input{sections/7conclusion}

\small{
\bibliographystyle{IEEE}
\bibliography{egbib}
}
\end{document}

%% file: sections/1intro.tex
\section{Introduction}
\label{sec:intro}
Airborne LiDAR is one of the most detailed and accurate methods for surveying large geographic areas quickly. It is able to capture topography data through vegetation and is more stable across illumination changes than photographic methods. Because of these benefits, along with the increasing availability of LiDAR sensors and the demand to produce high quality surveys, more organizations are relying on this technology. In addition to providing highly accurate point data, LiDAR also provides an intensity measurement. LiDAR intensity is recorded as the return strength or amplitude of the return signal. As intensity is directly related to surface reflectance and other surface characteristics, it has applications in feature extraction, classification, segmentation, surface analysis, object detection, and recognition \cite{intensityoverview}.
  
However, collecting airborne LiDAR over large areas can be very time consuming, and frequently requires multiple flights---either simultaneously or sequentially, to adequately map entire regions. This produces inconsistencies in the intensity measurement, as intensity itself is dependent upon the sensor's calibration, as well as on environmental factors, such as humidity, temperature, or wetness~\cite{intensityoverview}. By extension, intensities between adjacent or overlapping scans can be vastly different which is problematic when trying to use this measurement in many applications. 
    
Traditional methods for harmonizing the intensity involve multiple stages of processing. Starting from the raw intensity measurements, a correction model is typically employed to adjust the intensity values that reduces variation from the effects of various parameters such as range or angle of incidence. Secondly, most intensity processing systems utilize some normalization method that uses scaling or shifting to adjust the overall brightness to improve harmonization with neighboring tiles or overlapping regions~\cite{intensityoverview}. These processing methods can be difficult over the course of long collection campaigns. We propose a novel method for point cloud intensity harmonization using a deep neural network, which is capable of harmonizing scans from many different sources. We compare this method to several baselines, including interpolation-based methods as well as histogram matching. We show that our method is comparable to the best baseline in the simplest case, and surpasses it when there are distinct regions with unique physical brightness distributions present in the scan collection. Our method only requires the point cloud information from each scan and that there is sufficient overlap between scans.

%% file: sections/2related_works.tex
\section{Related Works}
\label{sec:Related Works}

\textbf{Radiometric Calibration} Scene radiance can be modeled as a nonlinear response function of the image brightness~\cite{kimthesis}. By using radiometric calibration, the brightness of the image can be mapped to a standardized unit, which makes it easier to compare images over a period of time. Radiometric calibration is used by many vendors working in airborne LiDAR to harmonize intensities, but this process is expensive. 

\textbf{Image Harmonization} There is significant research seeking to harmonize images for various contexts. For example, compositing is a technique that combines two or more images into a single image, often to create the illusion that the images are from the same scene. Several deep neural networks~\cite{qi2016pointnet, qi2017pointnet, luan2018deep} have been proposed to accomplish more seamless compositing between images. These methods harmonize composite images by translating the foreground into the domain of the background. Other methods~\cite{DBLP:colortransfer1, HWANG20191} seek only to harmonize the color between two images. A harmonization method for LiDAR would reduce or altogether eliminate the need for radiometric calibration. However, image harmonization techniques are not usable on point clouds. Unlike point clouds, images provide an inherent grid structure.

\textbf{Neural Network Models for Point clouds.} There have been many recent advancements in point cloud reasoning. The family of PointNet models~\cite{qi2016pointnet, qi2017pointnet} are designed to work on point clouds directly without imposing any additional structures. In contrast, some research has explored ways to understand point clouds using convolution, which is used to extract features from images. One example, KPConv~\cite{thomas2019kpconv} uses a deformable set of kernel points that act in much the same way as in image convolution. Other approaches explore transforming point clouds into more familiar structures. One such approach is Basis Point Sets~\cite{bps} which encodes a point cloud directly into a feature vector. In our work, we explore how a deep learning neural network might provide accurate color transfer across multiple scans.


%% file: sections/3problem_def.tex
\begin{figure}[t!]
    \centering
    \includegraphics[width=1\linewidth]{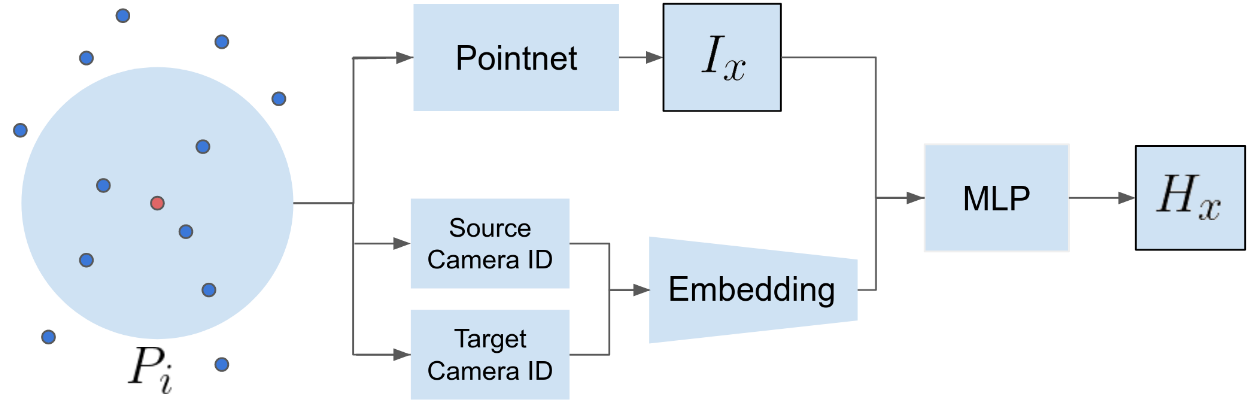}
    \caption{Our architecture. The intensity $I_x$ of neighborhood $P_i$ is interpolated at the target location (red). $I_x$ is concatenated with the difference of the source and target camera embeddings. It is then harmonized to the target scan's sensor calibration, $H_x$.}
    \label{fig:architecture}
\end{figure}

\section{LiDAR Harmonization}
\label{sec:LiDAR Harmonization}
We address the problem of harmonizing the intensity values of a set of overlapping point clouds, each of which is captured by sensors potentially featuring different models, calibrations, or capture conditions. Given a set of source clouds, we want to adjust the intensities of each point such that the source cloud intensity distribution matches the intensity distribution of the target cloud, but still conforms to the physical brightness distribution of the scan area. Similar to~\cite{kimthesis}, we define the relationship between a source point cloud intensity and the harmonized source point cloud intensity to be the nonlinear response function $I_x = f(H_x)$. $f$ represents the added linear and nonlinear differences that are inherent in the source sensor. We model this as a monotonic function, which is therefore invertible. Obtaining this inverse function provides a mapping from the source intensities to the target intensities, thereby harmonizing the source and target.
\begin{equation}
    \label{eqn:inverse}
    H_x = g(I_x), g = f^{-1}
\end{equation}
We propose a deep neural network based regression model capable of performing this task. Our network only relies on a sufficiently large overlap region between source and target scans. Our model is also not greatly affected by shifts in physical brightness distributions, such as transitions from dense urban areas to forested regions, both of which will have noticeably different intensity distributions.  

\subsection{Architecture}
An overview of our harmonization approach can be seen in \figref{architecture}. Given a point cloud neighborhood from a source scan with a central point $x$ and intensity $\hat{I_x}$, we want to predict the harmonized intensity of $x$ relative to a target scan, $\hat{H_x}$. In the ideal case, training would be as simple as finding source and target points with the same coordinate points and building $g$ from \eqnref{inverse} through any given function approximation algorithm. Having source-target point pairs with the same coordinates isolates the differences between sensors, and would make this task fairly trivial. However, it is rare for individual points from different scans to have the same exact coordinates. To address this, our architecture leverages a standard PointNet~\cite{qi2016pointnet} to accurately interpolate the intensity $I_x$ at the target point location within the source scan. Additionally, our architecture allows for harmonizing multiple sources by encoding the point source into a lower dimensional space. Specifically, we encode the point source into a 3 dimensional embedding with a dictionary size of 45. We learn this embedding lookup for each sensor. Our architecture then predicts the correct mapping $g$ for each source by passing the interpolated value concatenated with the difference between the source and target sensor embeddings into a multilayer perceptron (MLP), producing a harmonized value $H_x$. The loss for our model is formulated by the following equation, where $I_x$ is output of the PointNet, $H_x$ is the output of the MLP, and $\rho$ is the loss function:
\begin{flalign}
    \ell(\psi)_I &= \sum \rho(I_x, \hat{I_x}) \\
    \ell(\phi)_H &= \sum \rho(H_x, \hat{H_x}) \\
    L(\Theta)_T &= \ell(\psi)_I + \:\ell(\phi)_H
\end{flalign}
\subsection{Dataset} 
The NYU DublinCity LiDAR dataset~\cite{zolanvari2019dublincity} is a high resolution collection of 41 LiDAR scans over an area of Dublin, Ireland. These scans are already well harmonized, and so they provide a convenient ground truth by which to evaluate our method. 
From DublinCity we form a new dataset to train our method. After selecting a target scan $P_t$, all other scans with sufficient overlap in the target scan area are collected. The average scan size in DublinCity is around 30 million points. We define sufficient overlap to be at least 200 thousand points in the overlap region. Neighborhoods are then built by picking target points from $P_t$ and taking the closest 150 neighbors within 1 meter in the source scan. Using a set of monotonically increasing response functions from Columbia's Database of Response Functions~\cite{CAVE_0039}, we randomly assign response functions as synthetic corruption to each source scan. Examples are created from these source neighborhood-target point pairs, and corruption is applied to the neighborhoods based on their source. The target point is the harmonization ground truth, $\hat{H_x}$. The corruption transformation for each example is also applied to a copy of the target point and saved as the interpolation ground truth for that example, $\hat{I_x}$. 

In addition to these examples, we also sample points outside the overlap region for each scan. From these points, we build more neighborhoods as above. Since we are no longer in the overlap region, there is no target harmonization point. We utilize the same embedding since we are mapping within the same scan. While this seems like it should be a trivial operation, we found that it improved overall performance. We suspect this is because it improves the network's ability to interpolate. We apply the pre-assigned corruption to these neighborhoods as well. We define this collection of examples as well as the collection of examples from within the overlap region as the ``no shift" dataset.

DublinCity has a consistent intensity distribution for each scan. However, we wish to model regions that have physical brightness shifts, which are common in large LiDAR collections. A second dataset is created as an exact copy of the first. However, before applying the corruption, a global shift transformation is applied along the x-axis of the DublinCity LiDAR dataset. This transformation lowers the intensities over the left half of the region. This global shift simulates an area with physical brightness differences. In our implementation, we use a sigmoid transformation to achieve this effect: 
\begin{flalign*}
    \label{sigmoid}
    I_{x_{\text{shift}}} = s*I_xe^{-l(x-h)}+v
\end{flalign*}
where x is the normalized x-component of the point being shifted. This transformation produces a noticeable shift along x-axis with a transition zone that quickly ends the shift and returns to the original intensity distribution. In order to produce this significant shift in brightness, we use values $h=.5$, $v=.3$, $l=100$, and $s=.5$. We define this new dataset as the ``with shift" dataset. 

Stratified random sampling is used to balance the number of neighborhoods that come from the different source scans, as some scans have much larger overlap areas, and to balance the target intensities. In addition, we oversample the training dataset so that we have a balance of examples across the entire range of intensities for each scan.

\subsection{Implementation Details}
The Pytorch~\cite{pytorch} framework is used to implement and train our model. We use the Adam~\cite{adam} optimizer  with a cyclical learning~\cite{cyclical} rate, with maximum learning rate $1e^{-3}$ and minimum learning rate $1e^{-7}$. The learning rate is stepped up from the minimum to the maximum and back down each epoch. The maximum learning rate is lowered by 20 percent each epoch. We use a batch size of 50 and train for 40 epochs. Our MLP uses a hidden layer size of 100, with a ReLU activation layer and dropout, with dropout rate of 0.3.

%% file: sections/5evaluation.tex
\section{Evaluation}
\label{sec:Evaluation}
We evaluate our PointNet-interpolation method for LiDAR harmonization by comparing harmonized source scans to their original ground truth values. We report the mean absolute error (MAE) on two distinct datasets which are explained in the following section. We also compare our method to several baseline methods.

\subsection{Quantitative Analysis}
We evaluate our method as well as several baselines by comparing the harmonized output to the original ground truth values of each scan. Each baseline consists of one interpolation method paired with one harmonization method. The interpolation methods include standard interpolation, nearest neighbor interpolation, and cubic-spline interpolation. For harmonization, least squares approximation (``Linear") and the same MLP used in the PointNet-interpolation model are used. 

Finally, we compare our method with an entirely different approach: histogram matching. Histogram matching does not rely on the physical geometry of the point cloud but instead only depends on the entire distribution of intensities. Given a source and a reference distribution, histogram matching is able to transform the distribution to look like the reference. This technique is often used in image processing to balance contrast across an image by transforming the image's brightness distribution to be uniform. 

We evaluate our model and baselines by performing harmonization on the corrupted source scans. Since the scans are large, an evaluation tile is generated. This tile is chosen from an area that is outside the overlap region. We evaluate on numerous neighborhood sizes, but found that smaller neighborhood sizes were more effective. An overview of our results can be seen in \tblref{LiDARerror}. For all results, we use a neighborhood size of five. 

\subsection{Qualitative Analysis}
As can be seen in our table of results, our method performs exceedingly well on the ``With Shift" dataset versus all other baselines. We visualize the difference in performance between histogram matching and our method in \figref{qualitative}. The target scan is shown in (e), and the shifted region is visible on the left. The source scan (b) comes from this region, but histogram matching is biased towards the target distribution, as seen in (d).
Training requires a large number of training samples from across the range of intensities. Since we depend on what data from the only overlap region, it can be challenging to find intensities in certain ranges. This degrades our method's performance, which is noticeable in (c), as our model was unable to find an adequate sample of source neighborhoods with target pairs in the middle and upper ranges. 
\small{
\begin{figure}[t!]
    \centering
    \includegraphics[width=1\linewidth]{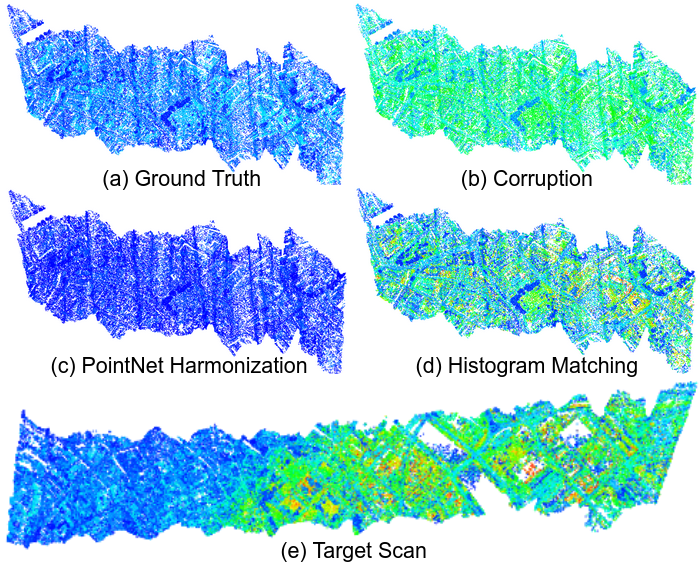}
        \caption{Qualitative results of our method compared to histogram matching. Color represents the intensity measurement. Our method is not affected by shifts in physical brightness distributions}
    \label{fig:qualitative}
\end{figure}}

\small{
\begin{table}[t]
    \centering
    {\renewcommand{\arraystretch}{1.2}%
    \begin{tabular}{|cc| c c |}
    \hline
    \multicolumn{2}{|c|}{Method}      & \multirow{2}{*}{No Shift} & \multirow{2}{*}{With Shift}\\
    Interpolation & Harmonization     &                                & \\\hline
    
    \multirow{2}{*}{Linear}  & MLP    & 0.077 & 0.073  \\
                             & Linear & 0.078 & 0.079  \\\hline
    \multirow{2}{*}{Cubic}   & MLP    & 0.073 & 0.075  \\
                             & Linear & 0.073 & 0.075  \\\hline
    \multirow{2}{*}{Nearest} & MLP    & 0.079 & 0.086  \\
                             & Linear & 0.073 & 0.090  \\\hline                          
                   PointNet  & MLP    & \textbf{0.052} & \textbf{0.040}  \\\hline
                  
    \multicolumn{2}{|c|}{Histogram Matching} & 0.053 & 0.138 \\\hline
    \end{tabular}}
    \caption{Quantitative harmonization results for different methods on DublinCity with and without a global shift. Results are given as mean absolute error (MAE).}
    \label{tab:LiDARerror}
\end{table}}

%% file: sections/7conclusion.tex
\section{Conclusion}
\label{sec:Conclusion}
We proposed an approach for LiDAR dataset harmonization that takes inspiration from approaches for image harmonization. The key challenges in the task are addressing the lack of truly matched pairs. We addressed this with a point cloud neural network architecture. Our approach is able to incorporate a variety of input features and is more accurate and robust than other approaches.

%% file: main.bbl
\begin{thebibliography}{10}

\bibitem{intensityoverview}
Alireza G.~Kashani, Michael Olsen, Christopher Parrish, and Nicholas Wilson,
\newblock ``A review of lidar radiometric processing: From ad hoc intensity
  correction to rigorous radiometric calibration,''
\newblock {\em Sensors}, vol. 15, pp. 28099--28128, 11 2015.

\bibitem{kimthesis}
Seon~Joo Kim,
\newblock {\em Radiometric Calibration Methods from Image Sequences},
\newblock Ph.D. thesis, University of North Carolina at Chapel Hill, USA, 2008.

\bibitem{qi2016pointnet}
Charles~R Qi, Hao Su, Kaichun Mo, and Leonidas~J Guibas,
\newblock ``{PointNet}: Deep learning on point sets for 3d classification and
  segmentation,''
\newblock in {\em IEEE Conference on Computer Vision and Pattern Recognition
  (CVPR)}, 2017.

\bibitem{qi2017pointnet}
Charles~Ruizhongtai Qi, Li~Yi, Hao Su, and Leonidas~J Guibas,
\newblock ``{PointNet++}: Deep hierarchical feature learning on point sets in a
  metric space,''
\newblock in {\em Advances in Neural Information Processing Systems (NeurIPS)},
  2017.

\bibitem{luan2018deep}
Fujun Luan, Sylvain Paris, Eli Shechtman, and Kavita Bala,
\newblock ``Deep painterly harmonization,'' 2018.

\bibitem{DBLP:colortransfer1}
Jianchao Tan, Jose~I. Echevarria, and Yotam~I. Gingold,
\newblock ``Palette-based image decomposition, harmonization, and color
  transfer,''
\newblock {\em CoRR}, vol. abs/1804.01225, 2018.

\bibitem{HWANG20191}
Youngbae Hwang, Joon-Young Lee, In~So Kweon, and Seon~Joo Kim,
\newblock ``Probabilistic moving least squares with spatial constraints for
  nonlinear color transfer between images,''
\newblock {\em Computer Vision and Image Understanding}, vol. 180, pp. 1 -- 12,
  2019.

\bibitem{thomas2019kpconv}
Hugues Thomas, Charles~R Qi, Jean-Emmanuel Deschaud, Beatriz Marcotegui,
  Fran{\c{c}}ois Goulette, and Leonidas~J Guibas,
\newblock ``{KPConv}: Flexible and deformable convolution for point clouds,''
\newblock in {\em IEEE International Conference on Computer Vision (ICCV)},
  2019.

\bibitem{bps}
Sergey Prokudin, Christoph Lassner, and Javier Romero,
\newblock ``Efficient learning on point clouds with basis point sets,''
\newblock in {\em IEEE International Conference on Computer Vision Workshops},
  2019.

\bibitem{zolanvari2019dublincity}
S.~M.~Iman Zolanvari, Susana Ruano, Aakanksha Rana, Alan Cummins,
  Rogerio~Eduardo da~Silva, Morteza Rahbar, and Aljosa Smolic,
\newblock ``{DublinCity}: Annotated lidar point cloud and its applications,''
\newblock {\em arXiv preprint arXiv:1909.03613}, 2019.

\bibitem{CAVE_0039}
M.D. Grossberg and S.K. Nayar,
\newblock ``What is the space of camera response functions?,''
\newblock in {\em IEEE Conference on Computer Vision and Pattern Recognition
  (CVPR)}, 2003.

\bibitem{pytorch}
Adam Paszke, Sam Gross, Soumith Chintala, Gregory Chanan, Edward Yang, Zachary
  DeVito, Zeming Lin, Alban Desmaison, Luca Antiga, and Adam Lerer,
\newblock ``Automatic differentiation in {PyTorch},''
\newblock in {\em NIPS Autodiff Workshop}, 2017.

\bibitem{adam}
Diederik~P. Kingma and Jimmy Ba,
\newblock ``Adam: A method for stochastic optimization,''
\newblock {\em CoRR}, vol. abs/1412.6980, 2014.

\bibitem{cyclical}
Leslie~N. Smith,
\newblock ``Cyclical learning rates for training neural networks,'' 2017.

\end{thebibliography}
